\documentclass[conference,a4paper]{IEEEtran}
\usepackage[pdftex]{graphicx}
\usepackage{multirow}
\usepackage{amsmath}
\usepackage{algorithmic}
\usepackage{array}
\usepackage{cite}
\usepackage{xcolor}
\usepackage{booktabs,ragged2e}
\usepackage[flushleft]{threeparttable}
\usepackage{cleveref}
\usepackage{stfloats}
\usepackage{pdfpages}

\newcommand{\regressor}[1]{{regressor}}

\usepackage{amsmath}
\usepackage{amsfonts}
\usepackage{amssymb}
\usepackage{amsthm}
\usepackage{bm}
\usepackage{bbm}
\usepackage{mathtools}
\usepackage{enumitem}
\usepackage{thmtools,thm-restate}
\usepackage[ruled,vlined]{algorithm2e}

\theoremstyle{plain}

\theoremstyle{definition}

\theoremstyle{remark}

\def\II{\mathcal{I}}
\def\LL{\mathcal{L}}
\def\NN{\mathcal{N}}

\def\UU{\mathcal{U}}
\def\VV{\mathcal{V}}

\def\Jb{\mathbf{J}}
\def\Mb{\mathbf{M}}

\def\rb{\mathbf{r}}

\def\xb{\mathbf{x}}
\def\yb{\mathbf{y}}\def\zb{\mathbf{z}}

\def\Ebb{\mathbb{E}}

\def\Rbb{\mathbb{R}}

\def\R{\Rbb}

\newcommand{\norm}[1]{ \| #1  \|  }

\DeclareMathOperator*{\argmin}{argmin}

\def\btheta{{\bm\theta}}
\def\bTheta{{\bm\Theta}}
\def\bbeta{{\bm\beta}}

\usepackage{xspace}
\makeatletter
\DeclareRobustCommand\onedot{\futurelet\@let@token\@onedot}
\def\@onedot{\ifx\@let@token.\else.\null\fi\xspace}

\def\etal{et al\onedot,}
\newcommand{\eg}{e.g., }
\newcommand{\ie}{i.e., }

\makeatother

\usepackage{etex,etoolbox}
\usepackage{environ}

\makeatletter

\begin{document}

\title{Semi-Supervised 3D Hand Shape and Pose Estimation with Label Propagation}

\author{\IEEEauthorblockN{Samira Kaviani, Amir Rahimi, Richard Hartley}
\IEEEauthorblockA{ANU, ACRV\\
\texttt{\{samira.kaviani,amir.rahimi,richard.hartley\}@anu.edu.au}
}
}

\maketitle
\IEEEpeerreviewmaketitle

\begin{abstract}
To obtain 3D annotations, we are restricted to controlled environments or synthetic datasets, leading us to 3D datasets with less generalizability to real-world scenarios.
To tackle this issue in the context of semi-supervised 3D hand shape and pose estimation, we propose the Pose Alignment network to propagate 3D annotations from labelled frames to nearby unlabelled frames in sparsely annotated videos.
We show that incorporating the alignment supervision on pairs of labelled-unlabelled frames allows us to improve the pose estimation accuracy.
Besides,  we show that the proposed Pose Alignment network can effectively propagate annotations on unseen sparsely labelled videos without fine-tuning.

\end{abstract}

\section{Introduction}\label{sec:introduction}
Estimating the 3D shape and pose of hands is an important problem in Computer Vision due to its fundamental role in various applications such as motion control~\cite{zhao2013robust}, 
human-computer interaction~\cite{yi2015atk}, and virtual/augmented reality~\cite{piumsomboon2013user}.

The performance of deep learning methods for 3D hand pose estimation remarkably depends on the quality of training data. 
Despite significant efforts in producing datasets with 3D ground-truth annotations, 
these datasets have limited generalizability to real-world scenarios.
The 3D ground-truth annotations are usually obtained in controlled environments using magnetic markers~\cite{garcia2018first}, multi-view settings~\cite{simon2017hand, zimmermann2019freihand, moon2020interhand2}, or synthetic datasets~\cite{hasson2019learning}.
Marker-based annotation approaches introduce biases in the trained models due to 
the visibility of the markers in the input images~\cite{Hasson_2020_CVPR}.
Synthetic datasets, on the other hand, suffer from the domain gap between natural and synthetic data.

Towards solving these challenges, in this paper, we address the problem of estimating the 3D hand pose and shape in a video dataset given only sparsely annotated frames. 
Motivated by the work of Bertasius~\etal~\cite{bertasius2019learning} for learning temporal human pose estimation in 2D, 
we propose the 3D Pose Alignment network where we propagate the annotation, consisting of hand root joint 3D location, MANO pose and shape parameters, from a labelled frame 
to a nearby unlabelled frame by learning the displacement of the two frames in parameter space. 
Our model consists of two components: 
\begin{enumerate}
\item A Single-frame Hand Predictor
that predicts  3D hand shape and pose from the input frames.
It also provides hand feature descriptors
from the input frames.
Similar to recent monocular RGB based hand predictors~\cite{hasson2019learning,Hasson_2020_CVPR},
a MANO layer~\cite{romero2017embodied}
is employed in our model to 
output the 3D hand joints and the corresponding mesh given the predicted 
hand shape and pose parameters.
\item  A Pose Alignment module that takes (i) the feature vectors
of an unlabelled and a nearby labelled frames, and 
(ii) the ground-truth annotations of the labelled frame as input and estimates the annotations of the unlabelled frame. 
\end{enumerate}
In this work, our goal is both to improve the Single-frame Hand Predictor using the information from frames without annotation and at the same time to learn the Pose Alignment network that can propagate annotations from a labelled frame to its nearby unlabelled frame.

We introduce a three-stage training scheme to train our model. 
(i) We first pre-train the Single-frame Hand Predictor and (ii) then Pose Alignment module on the labelled frames to provide a good initialization and informative hand features from the input frames.
(iii) In the last step of training, we perform training by passing 
pairs of frames consisting of an unlabelled frame $I_u$ and its nearby labelled frame $I_s$ from the same video to our model. 
We first compute the feature vectors
$\zb_s=\psi(I_s)$ and $\zb_u=\psi(I_u)$ corresponding to
labelled and unlabelled frames respectively and predict 
the annotations of the unlabelled frame,
which we represent by $\tilde{\gamma}_u$.
Then, the feature
vector difference between the labelled and unlabelled frames along with the predicted annotations $\tilde{\gamma}_u$ for the unlabelled frame are input to
the Pose Alignment module to compute a displacement required to
add to the unlabelled frame's predicted annotations
$\tilde{\gamma}_u$ to provide an estimate for the labelled
frame's annotations $\tilde{\gamma}_s =
\phi(\tilde{\gamma}_u, \zb_s - \zb_u) + \tilde{\gamma}_u$.

We define the loss on the output of the Pose Alignment module along with the output of the Hand Predictor on the labelled frames.
Our loss is defined by the difference between the estimated joints
and the ground-truth 3D joints annotation of the labelled frame. 

We evaluate our method under two different scenarios:
\begin{enumerate}
    \item  In the first scenario, we evaluate our Hand Predictor which is updated with the help of the Pose Alignment module on unlabelled frames to predict the 3D hand joints and the corresponding mesh given a single frame as input.
    
    \item In the second scenario, we evaluate our Pose Alignment module to propagate ground-truth annotations $\gamma^\star_s$ from a labelled frame $I_s$ to its nearby unlabelled frame $I_u$ within the same video, \ie $\tilde{\gamma}_u = \phi(\gamma^\star_s, \zb_u - \zb_s) + \gamma^\star_s $. 
 We show that our Pose Alignment module can provide high quality
 annotations for the unlabelled frames in an unseen video without any fine-tuning.
\end{enumerate}

\begin{figure*}[!ptb]
\begin{center}
\includegraphics[width=1.00\textwidth]{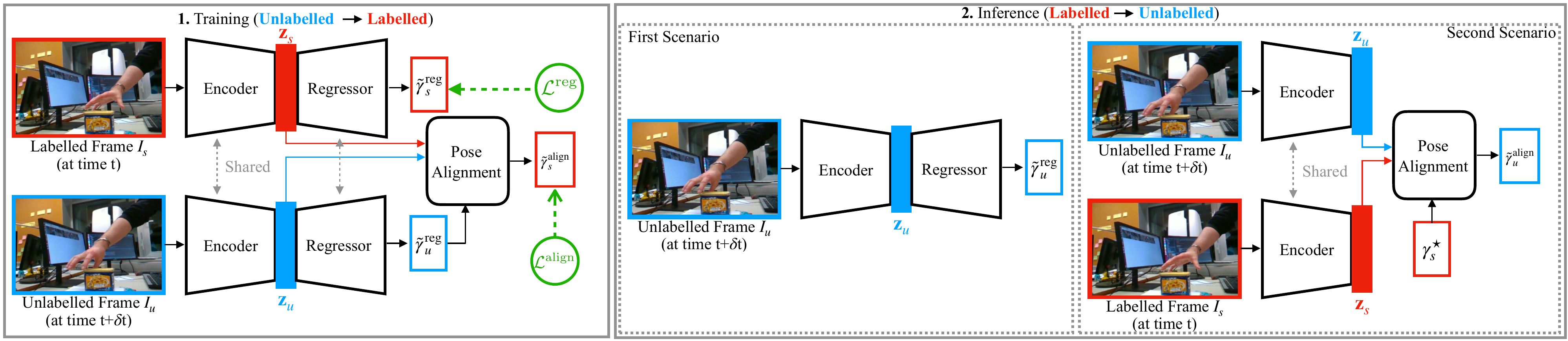}
\caption{\em Overview of our framework for semi-supervised 3D hand pose and
shape prediction. 
({\bf Left}) During training, we are given a labelled 
frame \textcolor{red}{$I_s$} at time $t$ and an 
unlabelled frame \textcolor{cyan}{$I_u$} at time $t+\delta t$ from
the same video. 
Both frames are passed through a shared encoder to obtain 
the corresponding features  $\textcolor{red}{\zb_s}$ and
$\textcolor{cyan}{\zb_u}$ for labelled and unlabelled frames, respectively. 
A \regressor{} is employed to predict the 3D hand shape and pose
parameters \textcolor{red}{$\tilde{\gamma}_s^{\rm reg}
$} 
and \textcolor{cyan}{$\tilde{\gamma}_u^{\rm reg}
$}. 
The predicted hand parameters \textcolor{cyan}{$\tilde{\gamma}_u^{\rm reg}$} 
of the unlabelled frame, along with the feature vectors \textcolor{red}{$\zb_s$},
and \textcolor{cyan}{$\zb_u$} are input to the Pose Alignment module to 
align the unlabelled frame hand parameters to the labelled frame hand 
parameters \textcolor{red}{$\tilde{\gamma}_s^{\rm align}$}. Our loss functions $\LL^{\rm reg}$ and $\LL^{\rm align}$ are defined on the outputs of both the \regressor{} and the Pose Alignment module. ({\bf Right}) We have two scenarios as our inference. In the first scenario (middle column), we use the output of the \regressor{} that is updated using the information from the unlabelled frames. In the second scenario (rightmost column), we use the Pose Alignment module to propagate annotations from the ground-truth parameters \textcolor{red}{$\gamma_s^\star$} of a labelled frame to predict the hand parameters \textcolor{cyan}{$\tilde{\gamma}_s^{\rm align}$} of a nearby unlabelled frame.
\label{Figure:overview}
}
\end{center}
\end{figure*}

The evaluations are performed on both unlabelled frames used during training and unlabelled frames of unseen test videos with sparse annotations.
Fig.~\ref{Figure:overview} illustrates our approach during training and inference.

\section{Method}\label{sec:method}
In this section, we first introduce our notation for 3D hand shape and pose estimation from monocular videos in the semi-supervised setting. Next, we describe each component of our model in detail. Finally, we explain the training, loss functions, and inference of our method.

\subsection{Notation} \label{sec:notation}
Let $\VV=\{V^i\}_{i=1}^{N}$ denote a set of $N$ video sequences, where each video sequence is an ordered set of frames denoted by $V^i=(I_1^i,\ldots,I_{T^i}^i)$.
Here, $T^i$ is the length of the video $V^i$, $I_t^i \in \II$ represents the $t$-th frame of video sequence $V^i$ and $\II$ is the input image space.
To simplify the notation, we may omit the $i$ superscript in the following sections and by $I_t$ we mean $t$-th frame of a random video sequence $V$ of length $T$.
We assume that the index set $\{1,\ldots,T\}$ for a video $V$ is decomposed into two disjoints sets, supervised~(labelled) frame indices $S_V$ and unsupervised~(unlabelled) frame indices $U_V$. 
Without loss of generality, we assume that the supervised frames are indexed by every $K$ frames in the video including the first and last frames, \ie $S_V=\{1,K,2K,\ldots,T\}$. 

In the semi-supervised setting, we have access to a set of  annotations $\{\gamma_s\mid s \in S_V\}$ for supervised frames of a video $V$.
In this problem, we use ``$\star$'' in superscript for ground-truth annotations. A ground-truth annotation
$\gamma^\star_s=(\rb^{\star}_s,\btheta^{\star}_s,\bbeta^{\star}_s)$ consists of $\rb^\star \in \R^3$ denoting the hand 3D root location in camera space, $\btheta^{\star}_s \in \R^{16\times3}$ denoting the axis-angle representation of joints rotation angles
(except fingertips), and $\bbeta^{\star}_s \in \R^{10}$ denoting the shape representation in the MANO model parametrization for the hand present in a frame $I_s$.
The hand parameters $\btheta^{\star}_s$ and $\bbeta^{\star}_s$ are fed to the MANO hand model $M(\btheta^{\star}_s, \bbeta^{\star}_s)$ to generate a triangulated 3D mesh $\Mb^{\star}_s \in \R^{778\times3}$ and its underlying 3D skeleton $\Jb^{\star}_s \in \R^{21\times3}$.

Given a video dataset $\VV$ with sparse annotations, our goal is to predict the hand 3D mesh and skeleton in the set of unsupervised frames $\UU=\bigcup_{V\in\VV}\{I_u \mid u \in U_V\}$.
We denote our predictions using the ``~$\tilde{}$~'' notation. 
As an example, 3D joint predictions for unsupervised frames are represented by $\tilde{\Jb}_u$. 
We assume we have access to the corresponding ground-truth joint coordinates ${\Jb}^\star_u$ of unsupervised frames for the evaluation purposes only.

\subsection{Components of the Model} \label{sec:comps}
\begin{figure}[!t]
\begin{center}
\includegraphics[width=\columnwidth]{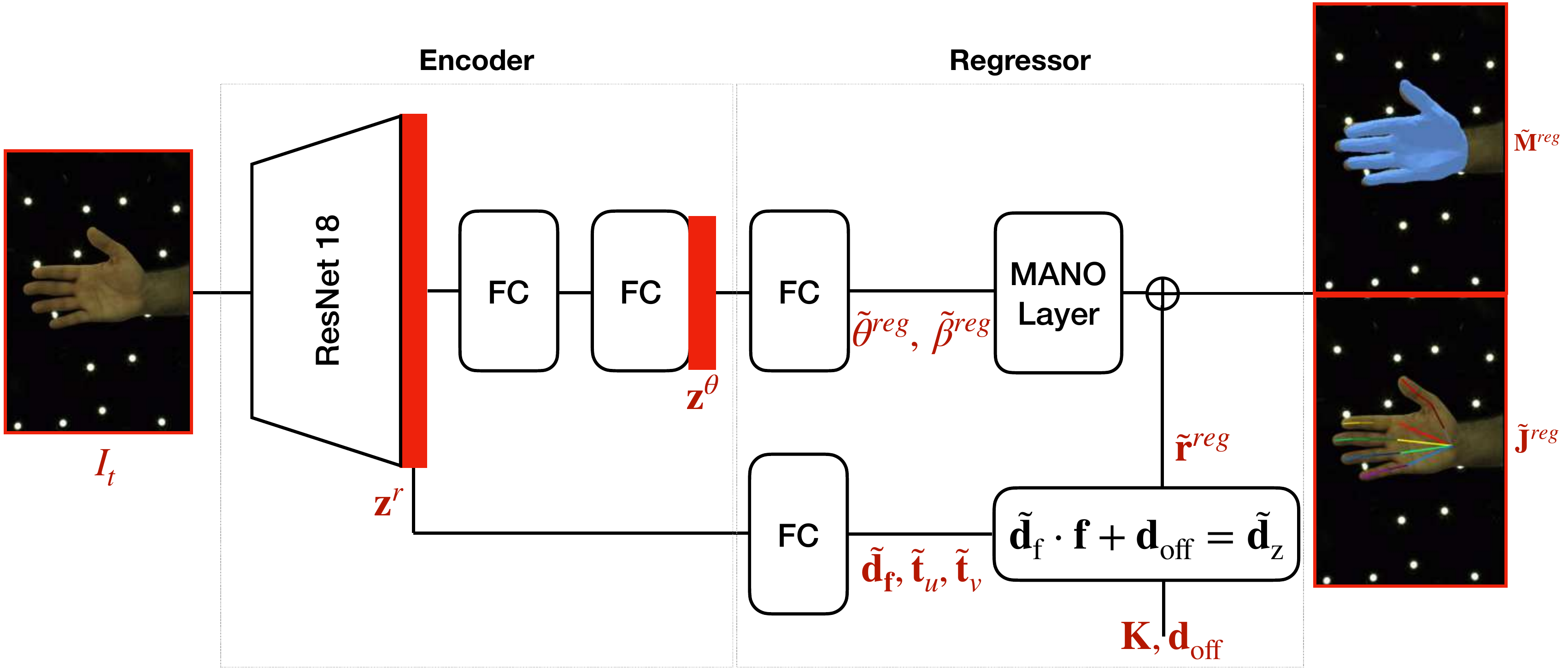}
\vspace{-0.75cm}
\caption{\em Hand Predictor. 
This module consists of an encoder and a \regressor{}.
The encoder receives an RGB image as input and extracts 
a low-level feature vector $\zb^r$ and a high-level feature vector $\zb^\theta$.
The \regressor{} takes the feature vectors $\zb^r,\zb^\theta$ and camera intrinsic parameters $\bf{K}$ as input and makes prediction for the MANO shape  $\tilde{\bbeta}^{\rm reg}$, pose  $\tilde{\btheta}^{\rm reg}$, and the hand 3D root location $\tilde{\rb}^{\rm reg}$ in camera space.
The estimated parameters of $\tilde{\bf d}_{\rm f}$ and $(\tilde{\bf t}_{\rm u}, \tilde{\bf t}_{\rm v})$ are the focal-normalized depth offset and the 2D translation vector in pixels with respect to the image center, respectively. Here,  $\bf{f}$  is the camera focal length and $\bf{d}_{\rm off}$ is empirically set for each dataset to obtain $\tilde{\bf d}_{\rm z}$ as the  depth distance between the hand root location and the camera center along the z-axis~\cite{Hasson_2020_CVPR}. %
\label{Figure:handpredictor}
}
\vspace{-0.5cm}
\end{center}
\end{figure}

\paragraph{Single-frame Hand Predictor}
In the Single-frame Hand Predictor module, the goal is to predict 3D mesh and skeleton of the hand present in an input frame $I$.
As shown in Fig.~\ref{Figure:handpredictor}, it consists of an encoder, and a \regressor{}. 

The encoder is represented by a function $\psi:\II \rightarrow \R^{2\times d}$, where the output feature vector $\zb=[\zb^r;\zb^\theta]=\psi(I)$ is composed of the concatenation of a low-level feature vector $\zb^r \in \R^d$ and a high-level feature vector $\zb^\theta \in \R^d$. 
The reason we use separate feature vectors in our model is that the low-level feature $\zb^r$ is better suited for representing the hand 3D root location $\rb \in \R^3$, while the high-level feature vector $\zb^\theta$ is more meaningful for predicting the MANO parameters~\cite{Hasson_2020_CVPR}. 

The \regressor{} takes the representation $\zb$ of a frame and predicts the MANO pose and shape parameters, and the hand 3D root location in camera space  $\tilde{\gamma}^{\rm reg}=(\tilde{\btheta}^{\rm reg}, \tilde{\bbeta}^{\rm reg}, \tilde{\rb}^{\rm reg})$.
We use the superscript ``${\rm reg}$'' to emphasize that these are outputs of the \regressor{}.
These predictions are computed by two separate branches $\pi^r: \R^d \rightarrow \R^3$ and $\pi^\theta: \R^d \rightarrow \R^{\left|{\btheta}\right| + \left|{\bbeta}\right|}$; one computing $\tilde{\rb}^{\rm reg}$ using $\zb^r$, and the other computing $\tilde{\btheta}^{\rm reg}$, and $\tilde{\bbeta}^{\rm reg}$ using $\zb^{\theta}$. Formally, we have
\begin{align}
    \tilde{\rb}^{\rm reg} &= \pi^r(\zb^r)~, \nonumber \\ 
    [\tilde{\btheta}^{\rm reg};\tilde{\bbeta}^{\rm reg}] &= \pi^\theta(\zb^\theta)~. \label{eq:reg}
\end{align}

\begin{figure}[!t]
\begin{center}
\includegraphics[width=\columnwidth]{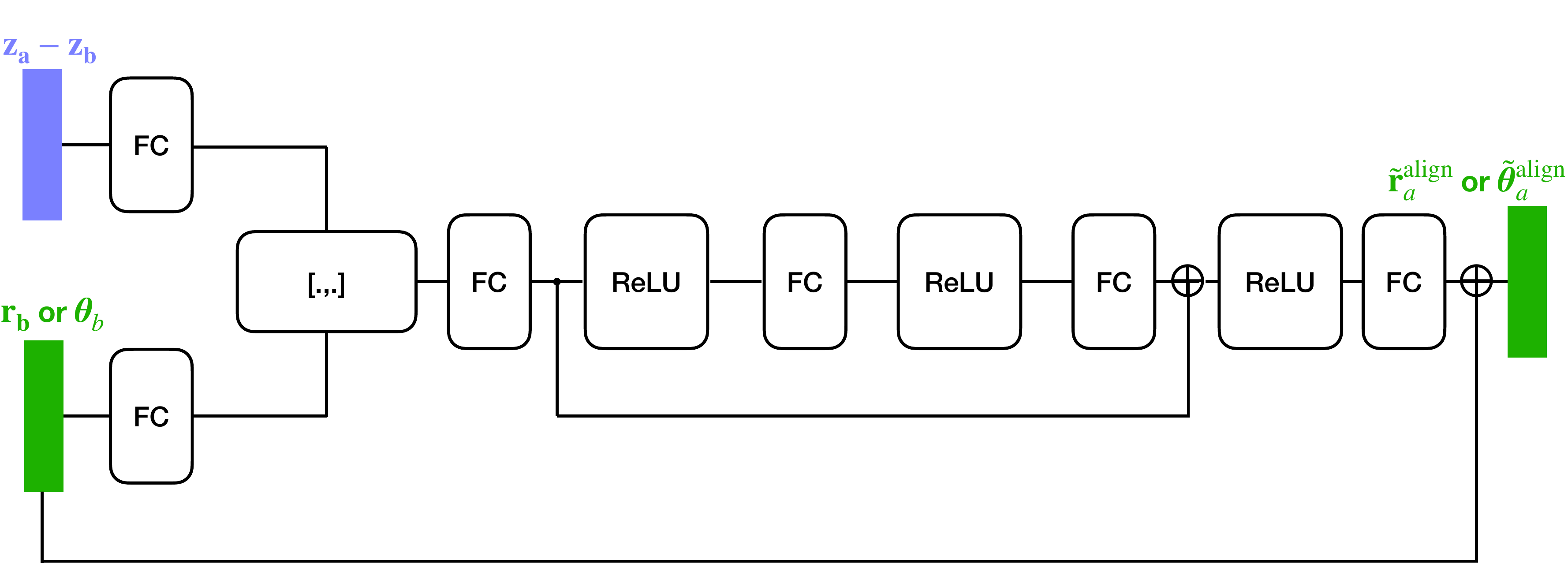}
\vspace{-0.7cm}
\caption{\em Pose Alignment.
This module takes the feature vectors $\zb_a$ and  $\zb_b$ extracted from the frames $I_a$ and $I_b$ and the hand parameters ($\rb_b$, or $ \btheta_b$ ) corresponding to the hand present in frame $I_b$ as input and predicts the hand parameters ($\rb_a$, or $ \btheta_a$) for the hand present in frame $I_a$.
\label{Figure:posealigner}
}
\end{center}
\vspace{-0.6cm}
\end{figure}

\paragraph{Pose Alignment}
This module receives feature vectors $\zb_a$ and $\zb_b$ from frames $I_a$ and $I_b$, and hand parameters $\gamma_b$
corresponding to the hand present in frame $I_b$ as input\footnote{Note that the input hand parameters can be the output of the Hand Predictor $\tilde{\gamma}^{\rm reg}_b$ or the ground-truth values $\tilde{\gamma}^{\star}_b$.
},
and predicts hand parameters
$\tilde{\gamma}^{\rm align}_a$ 
corresponding to the hand present in frame $I_a$.
Since the hands present in images $I_a$ and $I_b$ are assumed to be the same (\ie pairs of frames are sampled from the same monocular single-hand video), the value of the MANO shape parameter in frame $I_b$ is assigned to the predicted MANO shape parameter in frame $I_a$, that is,
\begin{equation}
    \tilde{\bbeta}^{\rm align}_a = \bbeta_b. 
    \label{eq:beta_align}
\end{equation}
To predict the hand root location in frame $I_a$, we have
\begin{equation}
    \tilde{\rb}^{\rm align}_a = \rb_b + \phi^{r}(\rb_b, \zb^{r}_{a}-\zb^r_{b})~,
    \label{eq:r_align}
\end{equation}
where $\phi^r: \R^3 \times \R^d \rightarrow \R^3$ is a function that computes a displacement required to add to $\rb_b$ given the difference of feature vectors, \ie $\zb^r_a-\zb^r_b$, to reach the location $\tilde{\rb}^{\rm align}_a$.
This function is implemented by a network composed of fully connected layers with residual connections as shown in Fig.~\ref{Figure:posealigner}.

For alignment in rotation space, we work with the axis-angle representation of joint rotations. 
To predict the hand joints rotations in frame $I_a$, we have
\begin{equation}
    \tilde{\btheta}^{\rm align}_a = {\btheta}_b + \phi^{\theta}({\btheta}_b, \zb^{\theta}_{a}-\zb^{\theta}_{b})~,
    \label{eq:theta_align}
\end{equation}
where $\phi^{\theta}: \R^{\left | \btheta \right| } \times \R^d \rightarrow \R^{\left | \btheta \right| }$ is a function~(see Fig.~\ref{Figure:posealigner}) that computes an offset required to add to $\btheta_b$ given the difference of feature vectors, \ie $\zb^{\theta}_a-\zb^{\theta}_b$, to reach  $\tilde{\btheta}^{\rm align}_a$.

\subsection{Training} \label{sec:training}
During training, our goal is to align the output of the Hand Predictor from an unsupervised frame $I_u$ to its nearest supervised frame $I_s$ with $s=\NN(u)=\argmin_{s' \in S_V} |s'-u|$. To this end, we define the following loss functions $\LL_{\rm {pred}}$ and $\LL_{\rm {align}}$.

\subsubsection{Single-frame Hand Predictor Loss}
The loss $\LL_{\rm {pred}}$ defines supervision on the output of the Hand Predictor given the supervised frames.
Similar to~\cite{Hasson_2020_CVPR}, it 
is defined as summation of $\LL_{\rm J}$, $\LL_{\theta}$ and $\LL_{\beta}$.
The loss $\LL_{\rm J}$ is defined on hand joints positions as squared $\ell_2$ distance.
Both $\LL_{\beta}$ and $\LL_{\theta}$ act as regularizations preventing unrealistic shape deformations. The losses are defined as
\begin{align}
    & \LL_{\rm J}(\Jb^{\star}_s, \tilde{\Jb}^{\rm reg}_s) = \norm{\Jb^{\star}_s - \tilde{\Jb}^{\rm reg}_s}^2_2~, \nonumber \\ 
    & \LL_{\theta}(\tilde{\btheta}^{\rm reg}_s) = \norm{\tilde{\btheta}^{\rm reg}_s}^2_2~, \nonumber \\ 
    & \LL_{\beta}(\tilde{\bbeta}^{\rm reg}_s) = \norm{\tilde{\bbeta}^{\rm reg}_s}^2_2~.
\end{align}
The Hand Predictor loss $\LL_{\rm pred}$ is defined as the total sum of these individual losses
\begin{equation}
    \LL_{\rm {pred}} = \lambda_{\rm J} \LL_{\rm J} + \lambda_{\theta} \LL_{\theta}+ \lambda_{\beta} \LL_{\beta}~,
\end{equation}
where $\lambda_J,\lambda_\theta, \lambda_\beta$ are scalar values.

\subsubsection{Pose Alignment Loss}
The loss $\LL_{\rm {align}}$ is defined on the output of the Pose Alignment module as
\begin{align}
    & \LL^{\rm align}_{\rm u \rightarrow s}(\Jb^{\star}_s, \tilde{\Jb}^{\rm align}_s) = \norm{\Jb^{\star}_s - \tilde{\Jb}^{\rm align}_s}^2_2~, \nonumber \\ 
    & \LL^{\rm align}_{\rm u \leftarrow s}(\tilde{\Jb}^{\rm reg}_u, \tilde{\Jb}^{\rm align}_u) = \norm{\tilde{\Jb}^{\rm reg}_u - \tilde{\Jb}^{\rm align}_u}^2_2~, \nonumber \\ 
    & \LL_{\theta}(\tilde{\btheta}^{\rm align}_s) = \norm{\tilde{\btheta}^{\rm align}_s}^2_2 + \norm{\tilde{\btheta}^{\rm align}_u}^2_2~,
\end{align}
where $\LL^{\rm align}_{u \rightarrow s}$ is the forward
alignment loss defined on the ground-truth annotation of
the supervised frame $I_s$, and the forward alignment from frame $I_u$ to $I_s$.
The loss $\LL^{\rm align}_{\rm u \leftarrow s}$
is the backward alignment loss defined on the output of the Hand
Predictor on the unsupervised frame $I_u$, and the backward
alignment from frame $I_s$ to $I_u$ using the Pose Alignment
module and $\LL_{\theta}$ is the regularization loss.
The backward alignment loss  $\LL^{\rm align}_{\rm u \leftarrow s}$ acts as a regularization enforcing the
output of the Hand Predictor for unsupervised frames to have
similar values to the output obtained by the Pose Alignment module.
The total alignment loss is defined as
\begin{equation}
    \LL_{\rm {align}} = \lambda_{\rm J} (  \LL^{\rm align}_{\rm u \rightarrow s} + \LL^{\rm align}_{\rm u \leftarrow s}) + \lambda_{\theta}  \LL_{\theta}~,
\end{equation}
with $\lambda_J,\lambda_\theta$ being scalar weights.\\

The total loss for a sampled frame $I_u$ and its nearest supervised frame $I_s$ is defined by
\begin{equation}
    \LL_{\rm total}(I_u, I_s) = 
    \LL_{\rm {pred}} +  \lambda_{\rm align} \LL_{\rm {align}}~,
    \label{eq:totalloss}
\end{equation}
with $\lambda_{\rm align}$ being a scalar value determining the importance of the Pose Alignment loss.
Denoting all the parameters in networks $\psi, \pi^r, \pi^\theta,\phi^r,\phi^\theta$ as $\bTheta$, our overall training objective  is to minimize the following function
\begin{equation}
    \argmin_{\bTheta} \mathop{\Ebb}_{V\sim\VV} \mathop{\Ebb}_{\substack{u\sim U_V\\s=\NN(u)}} \LL_{\rm total}(I_u, I_s)~.
\end{equation}
We remark that our goal is to not only train the Pose Alignment module, but also update the Hand Predictor with the help of the Pose Alignment module. These two modules cooperatively help each other to use the information from the unsupervised frames during training.

\subsection{Inference} \label{sec:inference}
During inference, we predict 3D hand shape and pose in an unsupervised frame $I_u$ by propagating the annotations from its nearby supervised frame $I_s$ (second scenario.)
To do so, we compute $(\tilde{\rb}^{\rm align}_u,\tilde{\btheta}^{\rm align}_u,\tilde{\bbeta}^{\rm align}_u)$ using the formulas in Eqs.~\ref{eq:beta_align},\ref{eq:r_align},\ref{eq:theta_align}. Since we update the Hand Predictor during training, we also compute its predictions $(\tilde{\rb}^{\rm reg}_u,\tilde{\btheta}^{\rm reg}_u,\tilde{\bbeta}^{\rm reg}_u)$ using Eq.~\ref{eq:reg} (first scenario.)

\section{Related Work}\label{sec:relatedwork}
\subsection{3D hand shape and pose from monocular RGB image}
There has been a surge of different deep learning approaches to predict hand shape and pose from monocular RGB images in recent years. These methods are broadly classified into model-based or model-free approaches.

Model-free approaches learn a mapping from image feature space to hand configuration space to predict only 3D hand skeletons without any reasoning about hand shape and surface~\cite{zimmermann2017learning, iqbal2018hand, cai2018weakly, spurr2018cross, yang2019disentangling, spurr2020weakly}.
For example, Zimmermann~\etal~\cite{zimmermann2017learning} propose a pipeline with three sequential stages of hand segmentation, 2D keypoints detection, and lifting the detected keypoints from 2D to 3D.
However, 3D keypoints predictions are only based on 2D keypoints predictions, and some image cues (\eg appearance, shading), which are helpful to reason depth, have been ignored.
In several other works, \eg~\cite{spurr2018cross, yang2019aligning, yang2019disentangling}, multimodal variational autoencoders~(VAE) are employed to predict a target modality given some other associated modalities (\eg RGB, 3D, and 2D hand skeletons).
Spurr~\etal~\cite{spurr2018cross} suggest learning a shared latent space between different modalities via a cross-modal training scheme. 
They embed each input modality to the shared latent space and make predictions either in the same or different modalities.
Yang~\etal~\cite{yang2019disentangling} introduce a disentangled cross-modal VAE to learn disentangled latent space of hand poses and hand images.
This approach allows better explicit controlling on the factors of variations (\eg pose, viewpoint, image background) for the hand image synthesis task. 
In another work, Yang~\etal~\cite{yang2019aligning} learn different latent spaces from different modalities jointly where the associated latent spaces are aligned via 
minimizing a $\rm KL$-divergence term.

Model-based approaches take advantage of an explicit deformable hand template that provides a strong inductive bias in learning the space of possible hand shapes and configurations. These approaches can predict 3D hand skeletons and their associated 3D surfaces by employing the hand templates in their predictions.
Panteleris~\etal~\cite{panteleris2018using} predict 3D hand shape as a post-processing step via fitting a 3D hand template to the detected 2D keypoints predicted by the opensource software OpenPose~\cite{simon2017hand}.
Some recent works~\cite{hasson2019learning, boukhayma20193d, baek2019pushing, zhang2019end, Hasson_2020_CVPR, liu2021semi, chen2021model} employ a differentiable hand deformation model known as MANO~\cite{romero2017embodied} in their framework to estimate 3D hand shape and pose jointly.
These methods usually are trained to predict MANO parameters (\ie pose and shape vectors) and a camera projection (\ie view parameters), given monocular RGB images. The estimated MANO parameters are fed into the MANO model to generate corresponding 3D hand meshes.
Different from these methods, some other works~\cite{ge20193d, kulon2019single, kulon2020weakly} predict hand meshes directly from image features.
These approaches try to predict hand meshes with either more resolution (\ie more mesh vertices) or better local details.
Ge~\etal~\cite{ge20193d} take the perspective that hand meshes are naturally graph-structured and propose a graph convolutional neural network to generate them.
Kulon~\etal~\cite{kulon2019single} train a graph convolutional autoencoder on a set of hand meshes. The decoder operates as a non-linear statistical morphable hand model taking the image latent code representation as input, generates the corresponding hand mesh.

\subsection{Limited data and supervision on 3D hand shape and pose from monocular RGB image}
Some prior works~\cite{cai2018weakly, ge20193d, boukhayma20193d, spurr2020weakly, kulon2020weakly, Hasson_2020_CVPR, liu2021semi, chen2021model}
address the issue of acquiring 3D annotations for in-the-wild datasets. This challenging problem requires multi-camera or motion capture studios.
For instance,~\cite{simon2017hand, zimmermann2019freihand, moon2020interhand2} are datasets captured in a calibrated multi-view setups and~\cite{garcia2018first} is a motion capture dataset in which magnetic sensors on hands are visible in images. 
While 3D annotation acquisition is complex, obtaining depth images from low-cost RGB-D cameras or 2D annotations, which can be detected or manually annotated on RGB images, is much easier.
Thus, to have a better generalization on in-the-wild images,
the community considered weakly supervised, semi-supervised, and self-supervised learning methods to efficiently take advantage of unlabelled data.

Both Cai~\etal~\cite{cai2018weakly} and Ge~\etal~\cite{ge20193d} train their networks on fully-annotated synthetic datasets combined with real-world datasets without 3D annotations by incorporating depth as a weak supervision signal.
Boukhayma~\etal~\cite{boukhayma20193d} notice that training with in-the-wild 2D annotations as weak supervision and full supervision on limited available datasets lead to more accurate results on in-the-wild images.
Spurr~\etal~\cite{spurr2020weakly} incorporate a set of biomechanical constraints as objective functions in training to penalize the anatomically invalid 3D predictions in a weakly-supervised scenario for a dataset with only 2D annotations.
Kolun~\etal~\cite{kulon2020weakly} generate 3D annotations (\ie 3D hand mesh and 3D keypoints) for unlabelled in-the-wild images from YouTube videos through fitting the MANO hand model to 2D keypoints detections provided by OpenPose.
They filter a subset of annotated samples based on the confidence scores obtained by OpenPose. 
Liu~\etal~\cite{liu2021semi} suggest a learning approach for annotating a large-scale unlabelled video dataset for hand-object pose estimation with the help of a fully annotated training dataset. They train their model on the fully annotated dataset and generate pseudo-annotations for the large-scale unlabelled dataset.
Then, they perform self-training on the subset of pseudo-annotations that are selected based on temporal consistency constraints.
Chen~\etal~\cite{chen2021model} utilize 2D keypoint detection as a weak form of supervisory signal in their framework to estimate textured 3D hand shape and pose from input images. They leverage geometric, photometric, and 2D-3D consistency as training objectives in their work.
Hasson~\etal~\cite{Hasson_2020_CVPR} extend the work in~\cite{hasson2019learning} to learn from sparsely-annotated videos.
Their intuition is that corresponding vertices in meshes from nearby frames should have the same color and propose exploiting this photometric consistency across adjacent frames as a supplementary cross-frame supervision.
Given a ground-truth mesh for a reference frame, an estimated mesh for a nearby frame, and camera intrinsic parameters, a 3D displacement (\ie flow) between the meshes is computed.
The nearby frame is warped into the reference frame using differentially rendered flow on the image plane.
Finally, photometric consistency between the warped and reference frames can be penalized as supervision.

Our work has a similar annotation setting to~\cite{Hasson_2020_CVPR}.
Despite that, our focus is to learn an effective pairwise alignment model (\ie motion model) to align hand pose annotations from a labelled frame to an unlabelled frame based on visual reasoning between them, under a semi-supervised regime. Instead of warping the frames in pixel space, our method employs a conditional displacement estimation between two meshes in the parameter space. The conditioning is defined based on the difference between the deep features extracted from the two images.
Unlike~\cite{hasson2019learning, Hasson_2020_CVPR}, we consider only
hand shape and pose estimation task and do not use any annotation about objects.

\section{Implementation}\label{sec:implementation}
The Single-frame Hand Predictor, referred to as {\em baseline}, 
is implemented similar to the hand-only branch of the single-frame hand-object 
network in~\cite{Hasson_2020_CVPR} adopting the same hyperparameters and objective
functions for training, unless otherwise noted.
It uses ResNet-18~\cite{he2016deep} as the backbone, where the extracted feature vectors from an RGB frame are $\zb^r \in \R^{512}$ and $\zb^\theta \in \R^{512}$.
Given the extracted feature vector $\zb=[\zb^r;\zb^\theta]$ and the camera intrinsic parameters, the \regressor{} outputs $\tilde{\gamma}^{\rm reg}=(\tilde{\rb}^{\rm reg} \in \R^3 ,\tilde{\bbeta}^{\rm reg} \in \R^{10},\tilde{\btheta}^{\rm reg} \in \R^{18})$, where
$\tilde{\btheta}^{\rm reg} \in \R^{3+15}$ represents the hand root rotation in the axis-angle representation and the $15$ PCA coefficients of the joints rotations.

The Pose Alignment module is implemented as shown in Fig.~\ref{Figure:posealigner}.
We use fully connected layers with $128$ hidden units followed by {ReLU} activation in the root aligner $\phi^r: \R^3 \times \R^{512} \rightarrow \R^3$.
A similar architecture with $512$ hidden units and {Tanh} as the last activation is employed for the rotation aligner $\phi^{\theta}: \R^{\left | \btheta \right| } \times \R^{512} \rightarrow \R^{\left | \btheta \right| }$.
Note that the rotation aligner works on the full $45$-dimensional joints rotations.
Therefore, we utilize the orthogonal basis matrix provided by the MANO layer to transform from the $15$-dimensional PCA space of the Hand Predictor into the full $45$-dimensional  joints rotations in axis-angle representation. These $45$-dimensional rotations together with the $3$-dimensional root joint rotation specify the $\left |\theta \right|=48$-dimensional rotations vector.

We train the model using the Adam optimizer~\cite{kingma2014adam} with a batch size of $64$ and a learning rate of $5\cdot10^{-5}$.
We assume that annotations for every $K=128$ frames, including the first and the last frames, are available in each video, \ie about $0.78125\%$ of frames are labelled.
We train our network in a three-stage procedure.
First, we pre-train the Single-frame Hand Predictor module on labelled frames with the loss $\LL_{\rm {pred}}$.
In the second stage, we only update the weights of the Pose Alignment module with the loss $\LL_{\rm {align}}$ on the samples from the labelled frame pairs.
In the last step, we train the whole network with the losses $\LL_{\rm {align}}$ and $\LL_{\rm {pred}}$ on pairs of frames selected as follows. We first randomly sample a labelled frame $I_s$. Then, the nearby frame $I_u$ is sampled from either labelled or unlabelled frames. When $I_u$ is sampled as an unlabelled frame, it is randomly selected from the frames with a gap less than $64$ time difference from $I_s$. When $I_u$ is sampled as a labelled frame, it is sampled from the top $3$ nearest labelled frames to $I_s$ including itself.
We set $\lambda_J=0.5$, $\lambda_\beta=5\cdot10^{-7}$, $\lambda_\theta=5\cdot10^{-6}$, and $\lambda_{\rm align}=0.1$.

\section{Experiments}\label{sec:experiments}
\subsection{Datasets}
\begin{figure*}[!ptb]
\label{Figure:qaulitative_results}
\begin{center}
\includegraphics[width=0.98\textwidth]{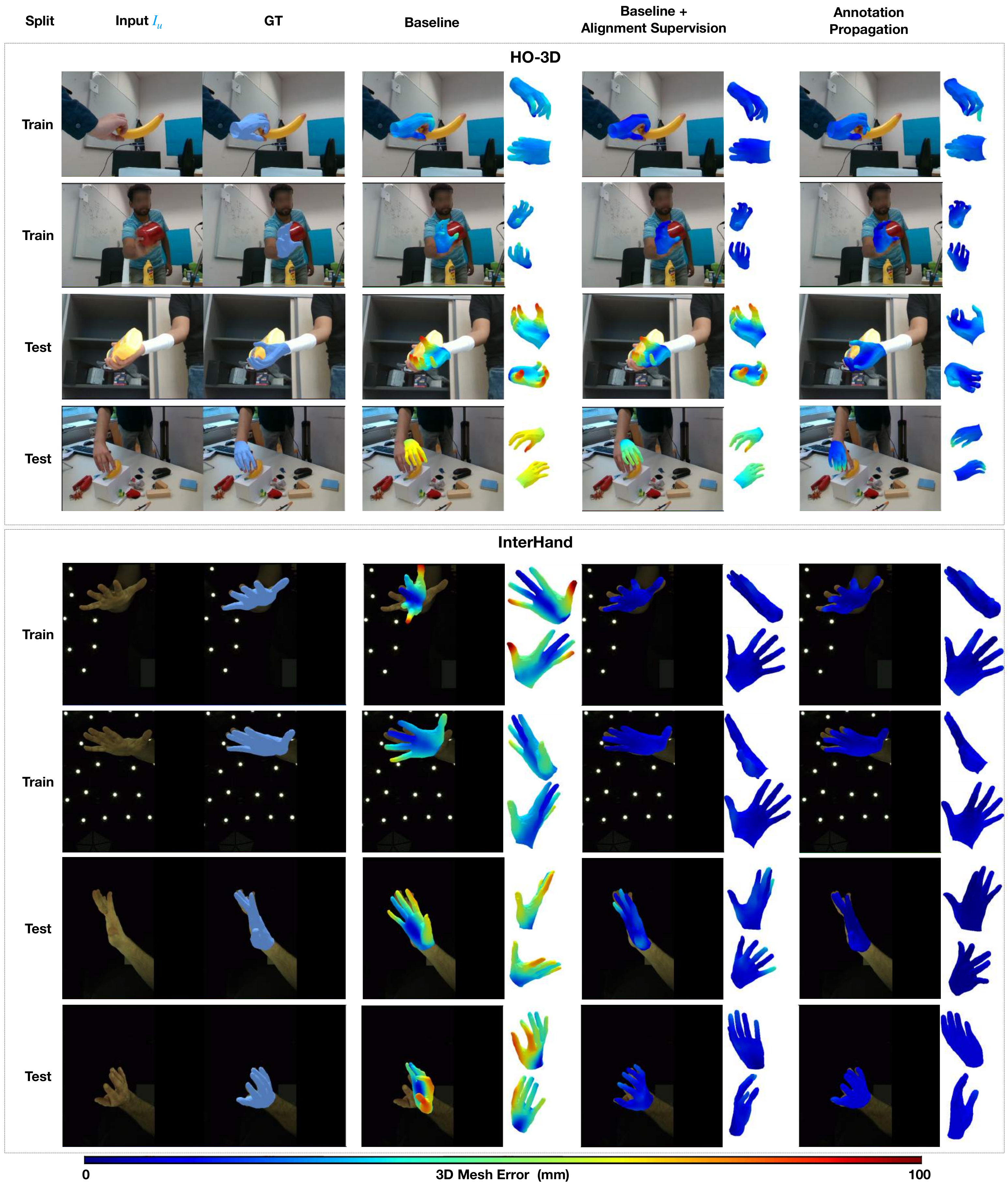}
\vspace{-0.2cm}
\caption{\em 
\label{Figure:qualitative_results}
Qualitative results of our annotation propagation and updated Hand Predictor~(Baseline+Alignment Supervision), compared to the baseline method on unlabelled frames in train and test splits of the Mini-HO-3D~({\bf top}) and Mini-InterHand~({\bf bottom}) datasets. The ground-truth and predicted meshes are rendered on the input images. The meshes are color-coded according to their error with respect to the corresponding ground-truth meshes using the color-bar shown at the bottom row. We illustrate the predicted hand meshes of different methods from two rotated views on the right hand side of the images.}
\end{center}
\vspace{-0.5cm}
\end{figure*}

We require long and diverse sequences to evaluate our method.
Thus, we consider two recently released datasets:

{\bf HO-3D~\cite{hampali2020honnotate}:} It is a 3D interacting hand-object pose dataset captured in a multi-RGB-D camera setup.
It consists of third-person view sequences with a right hand manipulating an object from the YCB dataset~\cite{xiang2017posecnn}.
In this work, we work on a subset of HO-3D, referred to as Mini-HO-3D containing $47k$ train, $4k$ val, and $14k$ test images.
Our specified subset is explained in more detail in the supplementary material.
{\bf InterHand2.6M~\cite{moon2020interhand2}:} It is a  3D interacting hand-hand pose dataset captured in a multi-view studio.
It includes sequences of a single hand or interacting hands with numerous poses, interactions, and camera views annotated semi-automatically.
We run experiments on a subset of the dataset, referred to as Mini-InterHand, containing  $319k$ train,  $48k$ val, and $144k$ test images.

\subsection{Evaluation Metrics}
\begin{table*}[!ptb]
\centering
\caption{\em \small 
Hand pose estimation performance on the unlabelled frames of the Mini-HO-3D and Mini-InterHand train and test sets using different evaluation metrics (in mm). Every $K=128$ frames in each video are annotated. The checkmark denotes the evaluation scenario. Our annotation propagation is denoted by \textcolor{blue}{Propagation} and is shown in \textcolor{blue}{blue}. Our updated Hand Predictor with the help of unlabelled data is denoted by Baseline+Align. 
\label{Table:quantitative_results}
}
\vspace{-0.2cm}
\resizebox{0.95\textwidth}{!}{%
\begin{tabular}{lllccccccccc}
\toprule
\multirow{2}{*}{Split} &
  \multicolumn{2}{l}{Scenario} &
  \multirow{2}{*}{Model} &
  \multirow{2}{*}{F@5$~\uparrow$} &
  \multirow{2}{*}{F@15$~\uparrow$} &
  \multirow{2}{*}{MPVPE$~\downarrow$} &
  \multirow{2}{*}{MPVPE~(P)$~\downarrow$} &
  \multirow{2}{*}{MPJPE$~\downarrow$} &
  \multirow{2}{*}{MPJPE~(T)$~\downarrow$} &
  \multirow{2}{*}{MPJPE~(SD)$~\downarrow$} &
  \multirow{2}{*}{MPJPE~(P)$~\downarrow$} \\
 &
  1st &
  2nd &
   &
   &
   &
   &
   &
   &
   &
   &
   \\
\midrule
\multicolumn{12}{c}{Mini-HO-3D}\\
\midrule
\multirow{3}{*}{Test} &
  \checkmark &
   &
  Baseline &
  \textbf{0.483} &
  \textbf{0.944} &
  64.813 &
  \textbf{10.688} &
  65.507 &
  \textbf{39.424} &
  46.351 &
  \textbf{10.892} \\
 &
  \checkmark &
   &
  Baseline + Align. &
  0.459 &
  0.940 &
  \textbf{62.337} &
  11.105 &
  \textbf{63.013} &
  40.169 &
  \textbf{45.275} &
  11.272 \\
 &
   &
  \checkmark &
  \textcolor{blue}{Propagation} &
  \textcolor{blue}{{\bf 0.677}} &
  \textcolor{blue}{{\bf 0.977}} &
  \textcolor{blue}{{\bf 43.083}} &
  \textcolor{blue}{{\bf 7.825}} &
  \textcolor{blue}{{\bf 44.014}} &
  \textcolor{blue}{{ 39.723}} &
  \textcolor{blue}{{\bf 38.426}} &
  \textcolor{blue}{{\bf 7.681}} \\
 \midrule
\multirow{3}{*}{Train} &
  \checkmark &
   &
  Baseline &
  \textbf{0.666} &
  \textbf{0.980} &
  \textbf{25.577} &
  \textbf{7.508} &
  \textbf{25.610} &
  \textbf{21.173} &
  \textbf{20.011} &
  \textbf{7.523} \\
 &
  \checkmark &
   &
  Baseline + Align. &
  0.662 &
  0.979 &
  25.891 &
  7.587 &
  25.957 &
  22.544 &
  20.487 &
  7.565 \\
 &
   &
  \checkmark &
  \textcolor{blue}{Propagation} &
  \textcolor{blue}{{\bf 0.666}} &
  \textcolor{blue}{{ 0.978}} &
  \textcolor{blue}{{ 27.169}} &
  \textcolor{blue}{{ 7.820}} &
  \textcolor{blue}{{ 27.232}} &
  \textcolor{blue}{{ 23.785}} &
  \textcolor{blue}{{ 21.766}} &
  \textcolor{blue}{{ 7.613}} \\
 \midrule
 \multicolumn{12}{c}{Mini-InterHand}\\
 \midrule
\multirow{3}{*}{Test} &
  \checkmark &
   &
  Baseline &
  0.457 &
  0.935 &
  40.349 &
  10.490 &
  41.282 &
  26.291 &
  21.555 &
  11.914 \\
 &
  \checkmark &
   &
  Baseline + Align. &
  \textbf{0.465} &
  \textbf{0.937} &
  \textbf{39.257} &
  \textbf{10.353} &
  \textbf{40.242} &
  \textbf{23.270} &
  \textbf{20.300} &
  \textbf{11.821} \\
 &
   &
  \checkmark &
  \textcolor{blue}{propagation} &
  \textcolor{blue}{{\bf 0.573}} &
  \textcolor{blue}{{\bf 0.938}} &
  \textcolor{blue}{{\bf 20.480}} &
  \textcolor{blue}{{\bf 9.646}} &
  \textcolor{blue}{{\bf 22.153}} &
  \textcolor{blue}{{\bf 19.981}} &
  \textcolor{blue}{{\bf 18.288}} &
  \textcolor{blue}{{ 11.837}} \\
\midrule
\multirow{3}{*}{Train} &
  \checkmark &
   &
  Baseline &
  0.573 &
  \textbf{0.958} &
  14.060 &
  8.382 &
  15.116 &
  13.981 &
  12.006 &
  \textbf{9.887} \\
 &
  \checkmark &
   &
  Baseline + Align. &
  \textbf{0.584} &
  \textbf{0.958} &
  \textbf{13.812} &
  \textbf{8.325} &
  \textbf{14.943} &
  \textbf{13.689} &
  \textbf{11.765} &
  9.908 \\
 &
   &
  \checkmark &
  \textcolor{blue}{propagation} &
  \textcolor{blue}{{\bf 0.666}} &
  \textcolor{blue}{{\bf 0.960}} &
  \textcolor{blue}{{\bf 13.552}} &
  \textcolor{blue}{{\bf 7.688}} &
  \textcolor{blue}{{ 15.393}} &
  \textcolor{blue}{{ 13.823}} &
  \textcolor{blue}{{ 12.383}} &
  \textcolor{blue}{{ 10.278}}\\
\bottomrule
\end{tabular}%
}
\vspace{-0.3cm}
\end{table*}

To evaluate 3D hand shape and pose methods, we measure the  {\em mean joints error}, {\em mean mesh error}, 
and {\em F-score} metrics. The evaluation metrics are described as follows:

{\bf Mean 3D error~\cite{zimmermann2017learning}:}
The {\em mean 3D error} (reported in mm) is defined as the average Euclidean distance between the corresponding predicted and ground-truth keypoints.
We denote {\em mean 3D joints error} as {\em MPJPE} (Mean Per Joint Position Error) and {\em mean 3D mesh error} as {\em MPVPE} (Mean Per Vertex Position Error).
Furthermore, as this task is an inherently ill-posed problem,
previous works, \eg \cite{zimmermann2019freihand, hampali2020honnotate}, suggest reporting the translation (T) aligned, and the Procrustes (P) aligned mean 3D errors to evaluate the articulation error.
The translation (T) aligned error is obtained after aligning the position of the predicted root keypoint with the ground-truth.
The Procrustes (P) aligned mean 3D error is obtained after Procrustes alignment (\ie overall scale, rotation, and mean position) of the predicted keypoints with ground-truth. In contrast to the previously proposed alignment methods, we believe that single image 3D hand pose estimation is ill-posed due to ambiguity in the depth and scale estimation (i.e., not the translation in $\xb$-$\yb$ coordinates or the 3D rotation).
Therefore, we propose the scaled-depth (SD) aligned mean 3D error, where the (SD) aligned error is obtained after aligning the overall scale and the mean depth of the predicted keypoints with the ground-truth.

{\bf F-score~\cite{knapitsch2017tanks}:}
It is a metric to measure how closely two point sets, with an unequal number of points, align to each other.
Given a distance threshold, {\em F-score} is defined as the harmonic mean of
recall and precision between two sets of points (\ie the predicted and the ground-truth points.) We use two distance thresholds as {\em F@5mm} and {\em F@15mm} to evaluate at fine and coarse scales following~\cite{zimmermann2019freihand}. We report the F-score values after Procrustes alignment of the points.

\subsection{Semi-supervised Evaluation}
We evaluate our method on the Mini-HO-3D and Mini-InterHand datasets.
We illustrate the effectiveness
of our approach on two different inference scenarios:
(i) improved pose estimation via incorporating alignment supervision, and (ii) annotation propagation.

{\bf Improved Pose Estimation via Incorporating Alignment Supervision:}
In the first scenario, we assess our method to
examine how well the proposed alignment supervision leads us
to take advantage of unlabelled frames during training.
We compare the performance of the {\em baseline} and the {\em baseline + alignment supervision}.
The {\em baseline} is the {\em Single-frame Hand Predictor}
trained with the loss $\LL_{\rm pred}$ on labelled frames of sparsely annotated train-set videos with similar hyperparameters as~\cite{Hasson_2020_CVPR}.
The {\em baseline + alignment supervision} has the same architecture as the {\em baseline}.
The difference is that its weights are updated with cooperation of the Pose Alignment module via the losses $\LL_{\rm pred}$ and $\LL_{\rm align}$ through training scheme mentioned in Sec.~\ref{sec:implementation}.

Table~\ref{Table:quantitative_results} presents the results on 
unlabelled frames of the train and test sets for the Mini-HO-3D and  Mini-InterHand datasets.
The {\em baseline + alignment supervision} on the Mini-HO-3D 
achieves an improvement of about $2.5$mm and $1.1$mm on unlabelled frames of the test-set for the {\em MPJPE} and {\em MPJPE (SD)} metrics, respectively.
The {\em baseline + alignment supervision } on the Mini-InterHand dataset yields improvement over most of the metrics for unlabelled frames of the train and test sets.
For example, it achieves an improvement of about $1$mm and $1.2$mm on unlabelled frames of the test-set for the {\em MPJPE} and {\em MPJPE (SD)} metrics, respectively.

{\bf Annotation Propagation:}
In the second scenario, we want to verify if
the Pose Alignment module learns to propagate information $\gamma^\star_s$ from manually annotated frames  $I_s$ to adjacent unlabelled frames $I_u$ within the same video, \ie $\tilde{\gamma}_u = \phi(\gamma^\star_s, \zb_u - \zb_s)$.
Table~\ref{Table:quantitative_results} also presents the accuracy of {\em annotation propagation} from a few manually labelled frames to unlabelled frames in the train and test sets of the Mini-HO-3D and Mini-InterHand datasets. 

The {\em annotation propagation} performance is in par with the {\em baseline} and {\em baseline + alignment supervision} for unlabelled frames in the train set of both the Mini-HO-3D and Mini-InterHand datasets.
The results obtained using our {\em annotation propagation} on unseen vidoes~(test set) show the superiority of our method
 against the {\em baseline } and {\em baseline + alignment} without any {\em fine-tuning}.

Fig.~\ref{Figure:qualitative_results} visualizes the qualitative results on the Mini-HO-3D and Mini-InterHand datasets comparing the {\em baseline}, {\em baseline + alignment supervision} and {\em annotation propagation} outputs on unlabelled samples from train and test splits.

\section{Conclusion}\label{sec:conclusion}
We introduce the Pose Alignment module to improve hand pose estimation 
via incorporating unlabeled data and defining the alignment objective functions.
Our experimental results demonstrate that our method improves hand pose estimation, 
and the proposed Pose Alignment module can effectively propagate annotations on 
unseen sparsely annotated videos without any fine-tuning.
In future works, we will consider resources of unlabelled in-the-wild video datasets
in training and exploring different alignment objectives, which can help us 
better generalize on real-world scenarios.

\vspace{0.2cm}
\noindent {\bf Acknowledgements.} {\small This research is supported in part by the Australia Research Council Centre of Excellence for Robotics Vision
(CE140100016).}

\bibliographystyle{IEEEtran}
\bibliography{Dicta_2021}

\begin{thebibliography}{10}
\providecommand{\url}[1]{#1}
\csname url@samestyle\endcsname
\providecommand{\newblock}{\relax}
\providecommand{\bibinfo}[2]{#2}
\providecommand{\BIBentrySTDinterwordspacing}{\spaceskip=0pt\relax}
\providecommand{\BIBentryALTinterwordstretchfactor}{4}
\providecommand{\BIBentryALTinterwordspacing}{\spaceskip=\fontdimen2\font plus
\BIBentryALTinterwordstretchfactor\fontdimen3\font minus
  \fontdimen4\font\relax}
\providecommand{\BIBforeignlanguage}[2]{{%
\expandafter\ifx\csname l@#1\endcsname\relax
\typeout{** WARNING: IEEEtran.bst: No hyphenation pattern has been}%
\typeout{** loaded for the language `#1'. Using the pattern for}%
\typeout{** the default language instead.}%
\else
\language=\csname l@#1\endcsname
\fi
#2}}
\providecommand{\BIBdecl}{\relax}
\BIBdecl

\bibitem{zhao2013robust}
W.~Zhao, J.~Zhang, J.~Min, and J.~Chai, ``Robust realtime physics-based motion
  control for human grasping,'' \emph{ACM Transactions on Graphics (TOG)},
  vol.~32, no.~6, pp. 1--12, 2013.

\bibitem{yi2015atk}
X.~Yi, C.~Yu, M.~Zhang, S.~Gao, K.~Sun, and Y.~Shi, ``Atk: Enabling ten-finger
  freehand typing in air based on 3d hand tracking data,'' in \emph{Proceedings
  of the 28th Annual ACM Symposium on User Interface Software \& Technology},
  2015, pp. 539--548.

\bibitem{piumsomboon2013user}
T.~Piumsomboon, A.~Clark, M.~Billinghurst, and A.~Cockburn, ``User-defined
  gestures for augmented reality,'' in \emph{IFIP Conference on Human-Computer
  Interaction}.\hskip 1em plus 0.5em minus 0.4em\relax Springer, 2013, pp.
  282--299.

\bibitem{garcia2018first}
G.~Garcia-Hernando, S.~Yuan, S.~Baek, and T.-K. Kim, ``First-person hand action
  benchmark with rgb-d videos and 3d hand pose annotations,'' in
  \emph{Proceedings of the IEEE conference on computer vision and pattern
  recognition}, 2018, pp. 409--419.

\bibitem{simon2017hand}
T.~Simon, H.~Joo, I.~Matthews, and Y.~Sheikh, ``Hand keypoint detection in
  single images using multiview bootstrapping,'' in \emph{Proceedings of the
  IEEE conference on Computer Vision and Pattern Recognition}, 2017, pp.
  1145--1153.

\bibitem{zimmermann2019freihand}
C.~Zimmermann, D.~Ceylan, J.~Yang, B.~Russell, M.~Argus, and T.~Brox,
  ``Freihand: A dataset for markerless capture of hand pose and shape from
  single rgb images,'' in \emph{Proceedings of the IEEE/CVF International
  Conference on Computer Vision}, 2019, pp. 813--822.

\bibitem{moon2020interhand2}
G.~Moon, S.-I. Yu, H.~Wen, T.~Shiratori, and K.~M. Lee, ``Interhand2. 6m: A
  dataset and baseline for 3d interacting hand pose estimation from a single
  rgb image,'' in \emph{Computer Vision--ECCV 2020: 16th European Conference,
  Glasgow, UK, August 23--28, 2020, Proceedings, Part XX 16}.\hskip 1em plus
  0.5em minus 0.4em\relax Springer, 2020, pp. 548--564.

\bibitem{hasson2019learning}
Y.~Hasson, G.~Varol, D.~Tzionas, I.~Kalevatykh, M.~J. Black, I.~Laptev, and
  C.~Schmid, ``Learning joint reconstruction of hands and manipulated
  objects,'' in \emph{Proceedings of the IEEE/CVF Conference on Computer Vision
  and Pattern Recognition}, 2019, pp. 11\,807--11\,816.

\bibitem{Hasson_2020_CVPR}
Y.~Hasson, B.~Tekin, F.~Bogo, I.~Laptev, M.~Pollefeys, and C.~Schmid,
  ``Leveraging photometric consistency over time for sparsely supervised
  hand-object reconstruction,'' in \emph{Proceedings of the IEEE/CVF Conference
  on Computer Vision and Pattern Recognition (CVPR)}, June 2020.

\bibitem{bertasius2019learning}
G.~Bertasius, C.~Feichtenhofer, D.~Tran, J.~Shi, and L.~Torresani, ``Learning
  temporal pose estimation from sparsely-labeled videos,'' \emph{Advances in
  Neural Information Processing Systems}, vol.~32, pp. 3027--3038, 2019.

\bibitem{romero2017embodied}
J.~Romero, D.~Tzionas, and M.~J. Black, ``Embodied hands: Modeling and
  capturing hands and bodies together,'' \emph{ACM Transactions on Graphics
  (ToG)}, vol.~36, no.~6, pp. 1--17, 2017.

\bibitem{zimmermann2017learning}
C.~Zimmermann and T.~Brox, ``Learning to estimate 3d hand pose from single rgb
  images,'' in \emph{Proceedings of the IEEE International Conference on
  Computer Vision}, 2017, pp. 4903--4911.

\bibitem{iqbal2018hand}
U.~Iqbal, P.~Molchanov, T.~B.~J. Gall, and J.~Kautz, ``Hand pose estimation via
  latent 2.5 d heatmap regression,'' in \emph{Proceedings of the European
  Conference on Computer Vision (ECCV)}, 2018, pp. 118--134.

\bibitem{cai2018weakly}
Y.~Cai, L.~Ge, J.~Cai, and J.~Yuan, ``Weakly-supervised 3d hand pose estimation
  from monocular rgb images,'' in \emph{Proceedings of the European Conference
  on Computer Vision (ECCV)}, 2018, pp. 666--682.

\bibitem{spurr2018cross}
A.~Spurr, J.~Song, S.~Park, and O.~Hilliges, ``Cross-modal deep variational
  hand pose estimation,'' in \emph{Proceedings of the IEEE Conference on
  Computer Vision and Pattern Recognition}, 2018, pp. 89--98.

\bibitem{yang2019disentangling}
L.~Yang and A.~Yao, ``Disentangling latent hands for image synthesis and pose
  estimation,'' in \emph{Proceedings of the IEEE Conference on Computer Vision
  and Pattern Recognition}, 2019, pp. 9877--9886.

\bibitem{spurr2020weakly}
A.~Spurr, U.~Iqbal, P.~Molchanov, O.~Hilliges, and J.~Kautz, ``Weakly
  supervised 3d hand pose estimation via biomechanical constraints,'' in
  \emph{Computer Vision--ECCV 2020: 16th European Conference, Glasgow, UK,
  August 23--28, 2020, Proceedings, Part XVII 16}.\hskip 1em plus 0.5em minus
  0.4em\relax Springer, 2020, pp. 211--228.

\bibitem{yang2019aligning}
L.~Yang, S.~Li, D.~Lee, and A.~Yao, ``Aligning latent spaces for 3d hand pose
  estimation,'' in \emph{Proceedings of the IEEE International Conference on
  Computer Vision}, 2019, pp. 2335--2343.

\bibitem{panteleris2018using}
P.~Panteleris, I.~Oikonomidis, and A.~Argyros, ``Using a single rgb frame for
  real time 3d hand pose estimation in the wild,'' in \emph{2018 IEEE Winter
  Conference on Applications of Computer Vision (WACV)}.\hskip 1em plus 0.5em
  minus 0.4em\relax IEEE, 2018, pp. 436--445.

\bibitem{boukhayma20193d}
A.~Boukhayma, R.~d. Bem, and P.~H. Torr, ``3d hand shape and pose from images
  in the wild,'' in \emph{Proceedings of the IEEE/CVF Conference on Computer
  Vision and Pattern Recognition}, 2019, pp. 10\,843--10\,852.

\bibitem{baek2019pushing}
S.~Baek, K.~I. Kim, and T.-K. Kim, ``Pushing the envelope for rgb-based dense
  3d hand pose estimation via neural rendering,'' in \emph{Proceedings of the
  IEEE/CVF Conference on Computer Vision and Pattern Recognition}, 2019, pp.
  1067--1076.

\bibitem{zhang2019end}
X.~Zhang, Q.~Li, H.~Mo, W.~Zhang, and W.~Zheng, ``End-to-end hand mesh recovery
  from a monocular rgb image,'' in \emph{Proceedings of the IEEE/CVF
  International Conference on Computer Vision}, 2019, pp. 2354--2364.

\bibitem{liu2021semi}
S.~Liu, H.~Jiang, J.~Xu, S.~Liu, and X.~Wang, ``Semi-supervised 3d hand-object
  poses estimation with interactions in time,'' in \emph{Proceedings of the
  IEEE/CVF Conference on Computer Vision and Pattern Recognition}, 2021, pp.
  14\,687--14\,697.

\bibitem{chen2021model}
Y.~Chen, Z.~Tu, D.~Kang, L.~Bao, Y.~Zhang, X.~Zhe, R.~Chen, and J.~Yuan,
  ``Model-based 3d hand reconstruction via self-supervised learning,'' in
  \emph{Proceedings of the IEEE/CVF Conference on Computer Vision and Pattern
  Recognition}, 2021, pp. 10\,451--10\,460.

\bibitem{ge20193d}
L.~Ge, Z.~Ren, Y.~Li, Z.~Xue, Y.~Wang, J.~Cai, and J.~Yuan, ``3d hand shape and
  pose estimation from a single rgb image,'' in \emph{Proceedings of the
  IEEE/CVF Conference on Computer Vision and Pattern Recognition}, 2019, pp.
  10\,833--10\,842.

\bibitem{kulon2019single}
D.~Kulon, H.~Wang, R.~A. G{\"{u}}ler, M.~M. Bronstein, and S.~Zafeiriou,
  ``Single image 3d hand reconstruction with mesh convolutions,'' in
  \emph{Proceedings of the British Machine Vision Conference ({BMVC})}, 2019.

\bibitem{kulon2020weakly}
D.~Kulon, R.~A. Guler, I.~Kokkinos, M.~M. Bronstein, and S.~Zafeiriou,
  ``Weakly-supervised mesh-convolutional hand reconstruction in the wild,'' in
  \emph{Proceedings of the IEEE/CVF Conference on Computer Vision and Pattern
  Recognition}, 2020, pp. 4990--5000.

\bibitem{he2016deep}
K.~He, X.~Zhang, S.~Ren, and J.~Sun, ``Deep residual learning for image
  recognition,'' in \emph{Proceedings of the IEEE conference on computer vision
  and pattern recognition}, 2016, pp. 770--778.

\bibitem{kingma2014adam}
D.~P. Kingma and J.~Ba, ``Adam: A method for stochastic optimization,''
  \emph{arXiv preprint arXiv:1412.6980}, 2014.

\bibitem{hampali2020honnotate}
S.~Hampali, M.~Rad, M.~Oberweger, and V.~Lepetit, ``Honnotate: A method for 3d
  annotation of hand and object poses,'' in \emph{Proceedings of the IEEE/CVF
  Conference on Computer Vision and Pattern Recognition}, 2020, pp. 3196--3206.

\bibitem{xiang2017posecnn}
Y.~Xiang, T.~Schmidt, V.~Narayanan, and D.~Fox, ``Posecnn: A convolutional
  neural network for 6d object pose estimation in cluttered scenes,''
  \emph{Robotics: Science and Systems (RSS)}, 2018.

\bibitem{knapitsch2017tanks}
A.~Knapitsch, J.~Park, Q.-Y. Zhou, and V.~Koltun, ``Tanks and temples:
  Benchmarking large-scale scene reconstruction,'' \emph{ACM Transactions on
  Graphics (ToG)}, vol.~36, no.~4, pp. 1--13, 2017.

\end{thebibliography}

\newpage
\appendices
\section{Datasets Subset}\label{appendix:datasets_subset}

{\bf HO-3D Subset:}
To form the Mini-HO-3D  dataset,
out of train sequences,
we selected sequences with names of {\em \small{ ABF11, BB12, GPMF13, GSF14}} as val split and with names of {\em \small{ABF10, MC1, MDF14, BB11, GPMF12, GSF13, SB14, ShSu10, SM3, SMu1, SiBF11, SS2}} as test split and the rest sequences as train split.

{\bf InterHand Subset:}
To form the Mini-InterHand dataset, we need long sequences with only the right hand.
The InterHand train set sequences with only the right hand have a maximum length of $97$ frames. Thus,
We filtered out sequences with the right hand and a minimum length of $1000$ frame from the val and test sets.
Out of the filtered val split, we selected the last $25$ videos as our val-set and the rest of that as our train-set.
Out of the filtered test split, we randomly selected $100$ videos as our test-set.

\section{More samples from our experiments} \label{appendix:more_samples}
\begin{figure*}[!ptb]
\begin{center}
\includegraphics[width=1.00\textwidth]{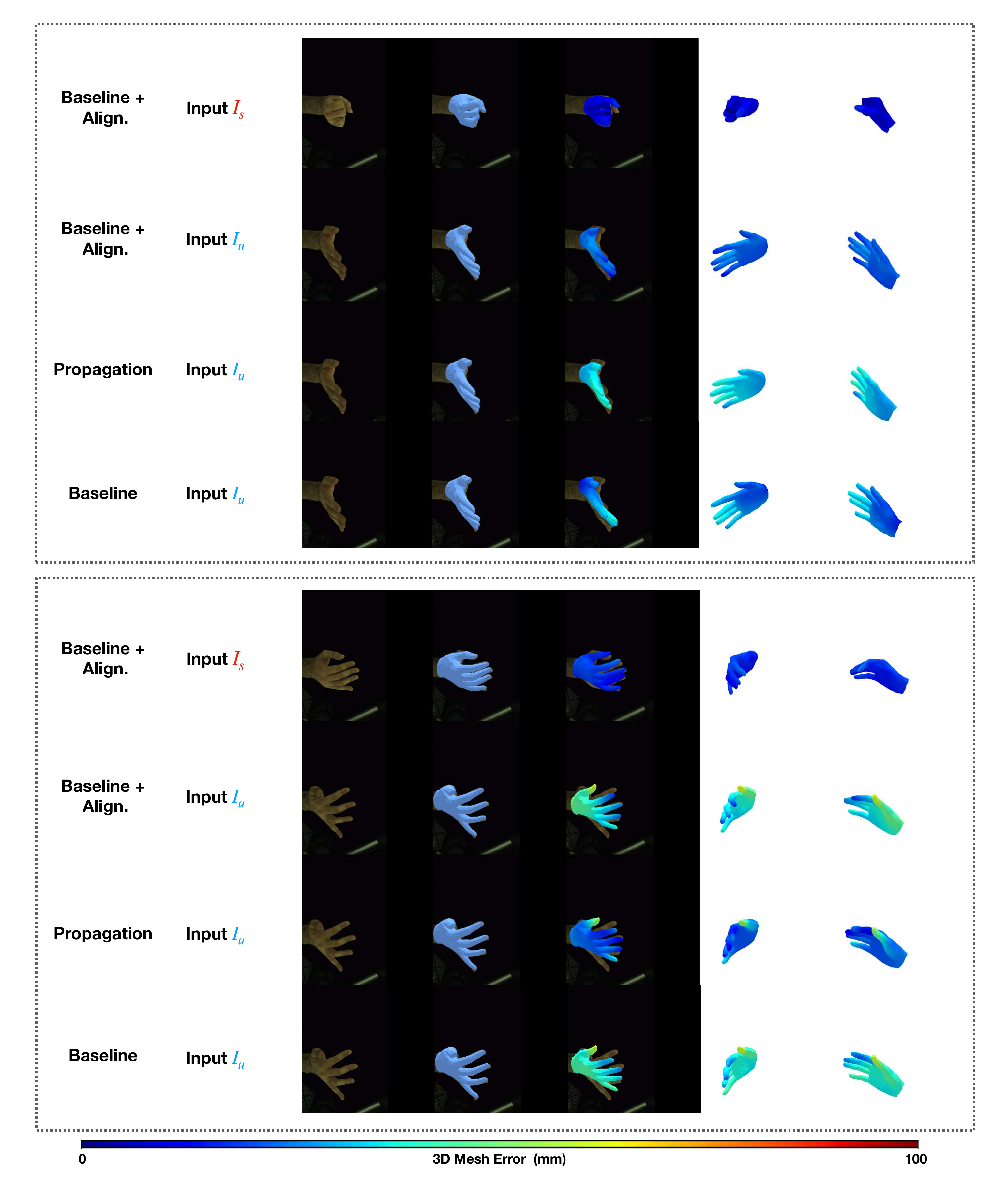}
\caption{\em 
Qualitative results on pair labelled \textcolor{red}{$I_s$} and unlabelled \textcolor{cyan}{$I_u$} frames in val split of the Mini-InterHand.
\label{Figure:a1}
}
\end{center}
\end{figure*}

\begin{figure*}[!ptb]
\begin{center}
\includegraphics[width=1.00\textwidth]{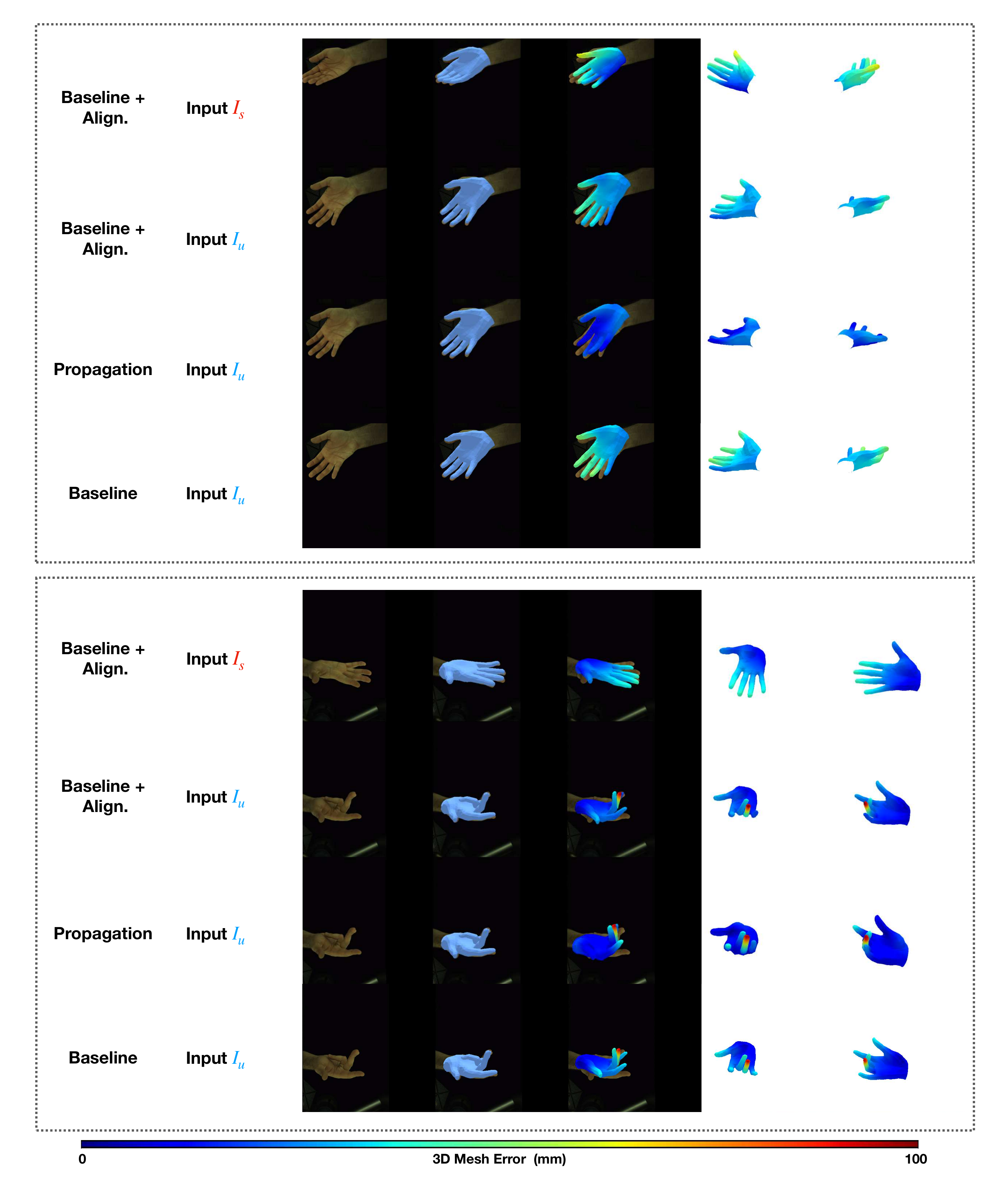}
\caption{\em 
Qualitative results on pair labelled \textcolor{red}{$I_s$} and unlabelled \textcolor{cyan}{$I_u$} frames in val split of the Mini-InterHand.
\label{Figure:a2}
}
\end{center}
\end{figure*}

\begin{figure*}[!ptb]
\begin{center}
\includegraphics[width=1.00\textwidth]{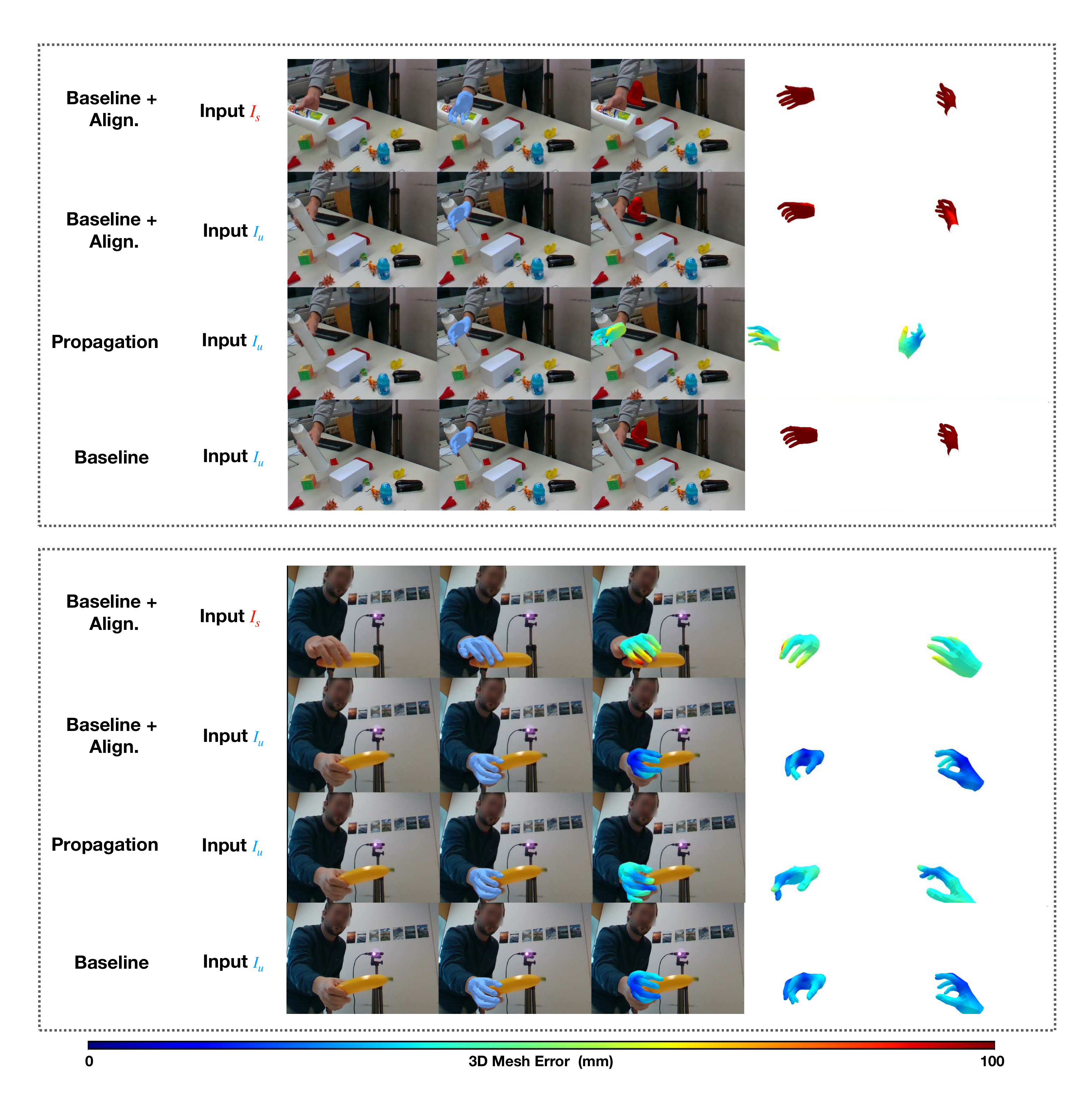}
\caption{\em 
Qualitative results on pair labelled \textcolor{red}{$I_s$} and unlabelled \textcolor{cyan}{$I_u$} frames in val split of the Mini-HO-3D.
\label{Figure:a3}
}
\end{center}
\end{figure*}

\begin{figure*}[!ptb]
\begin{center}
\includegraphics[width=1.00\textwidth]{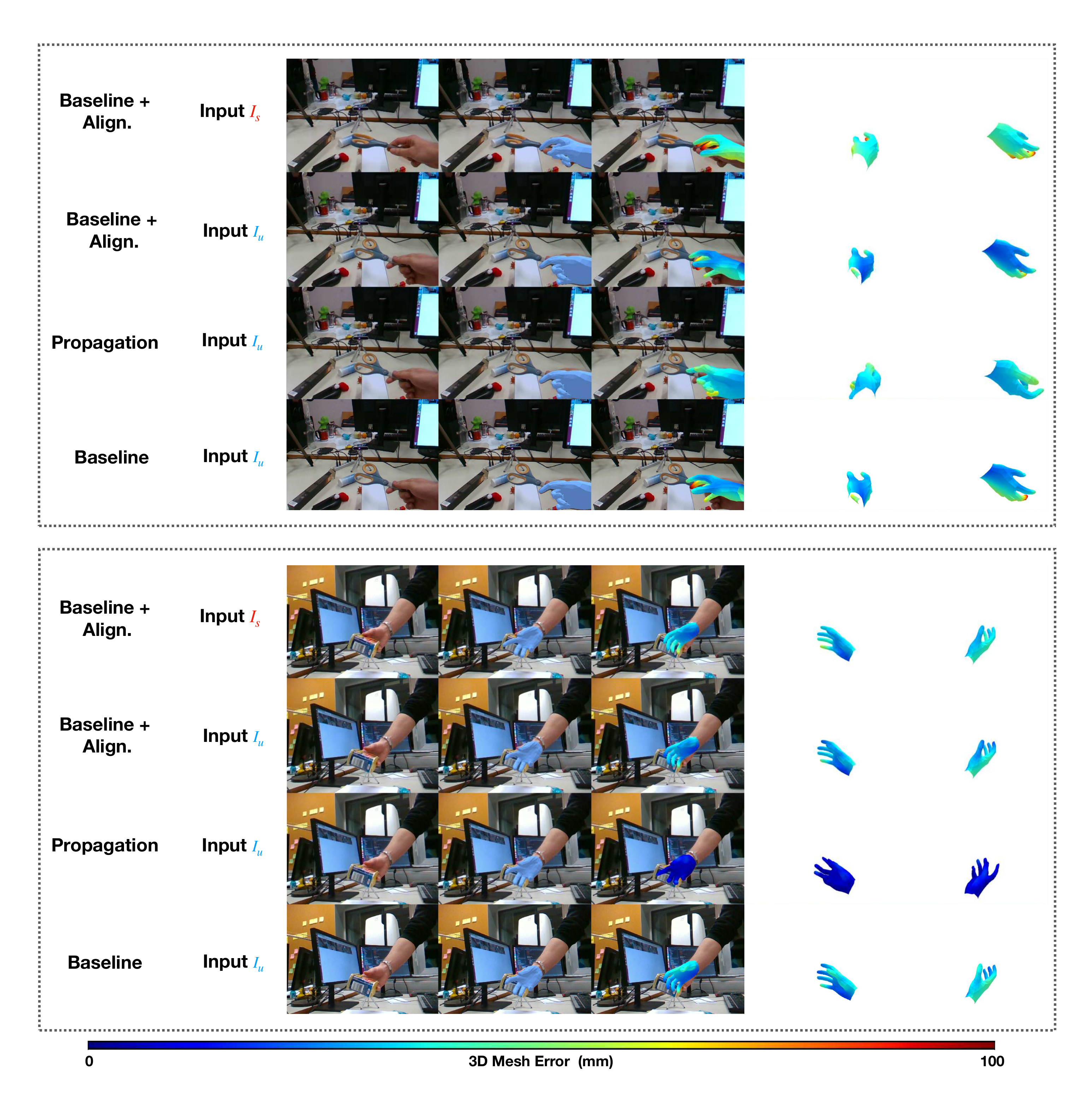}
\caption{\em 
Qualitative results on pair labelled \textcolor{red}{$I_s$} and unlabelled \textcolor{cyan}{$I_u$} frames in val split of the Mini-HO-3D.
\label{Figure:a4}
}
\end{center}
\end{figure*}

Fig.~\ref{Figure:a1},~\ref{Figure:a2},~\ref{Figure:a3},~\ref{Figure:a4} show qualitative results of our {\em annotation propagation} and {\em baseline + alignment supervision}, compared to the {\em baseline} method on pairs of labelled \textcolor{red}{$I_s$} and unlabelled \textcolor{cyan}{$I_u$} frames sampled from val splits of the Mini-InterHand and Mini-HO-3D datasets.
Every $K=128$ frames in each video are annotated.
The {\em baseline} and {\em baseline + alignment supervision} methods are evaluated following the first scenario and the {\em annotation propagation} method is evaluated following the second scenario.

\begin{table*}[!hpb]
\centering
\caption{\em \small 
Hand pose estimation performance compared with state-of-the-art method on all frames of test split of Mini-HO-3D dataset. 
The checkmarks show that the methods include 6D object pose estimation.
\label{Table:sota_compare}
}
\vspace{-0.2cm}
\resizebox{0.95\textwidth}{!}{%
\begin{tabular}{cccccccccc}
\toprule
Model &
  F@5$~\uparrow$ &
  F@15$~\uparrow$ &
  MPVPE$~\downarrow$ &
  MPVPE~(P)$~\downarrow$ &
  MPJPE$~\downarrow$ &
  MPJPE~(T)$~\downarrow$ &
  MPJPE~(SD)$~\downarrow$ &
  MPJPE~(P)$~\downarrow$ &
  Object Estimation\\
\midrule
Baseline &
  \textbf{0.483} &
  \textbf{0.944} &
  64.818 &
  \textbf{10.690} &
  65.513 &
  \textbf{39.418} &
  46.351 &
  \textbf{10.893} &
   \\
Baseline + Align. &
  0.458 &
  0.940 &
  \textbf{62.341} &
  11.107 &
  \textbf{63.018} &
  40.158 &
  \textbf{45.273} &
  11.274 &
   \\
\midrule
Hasson~\etal~\cite{Hasson_2020_CVPR}~Hand-Object Net &
  0.336 &
  0.915 &
  73.448 &
  16.144 &
  73.517 &
  48.449 
  &
  54.550 &
  17.844 &
  \checkmark
   \\
Hasson~\etal~\cite{Hasson_2020_CVPR}~Hand-Object Net + Consist. &
  0.330 &
  0.850 &
  80.739 &
  15.108 &
  80.001 &
  60.517 
  &
  53.581 &
  15.951 &
  \checkmark \\
\bottomrule
\end{tabular}%
}
\vspace{-0.3cm}
\end{table*}

\section{Comparisons with State-of-the-arts} \label{appendix:compare_w_sota}
We compare our method with the state-of-the-art method~\cite{Hasson_2020_CVPR}, retrained on the Mini-HO-3D dataset. We use their provided source code and follow their instruction for training the model.
Following the training procedure and hyperparameters in~\cite{Hasson_2020_CVPR},
we first pretrain the {\em single-frame hand-object network} on fractions of the data (\ie every $K=128$ frames) without the consistency loss for
$53k$ iterations with the batch size of $8$.
Then, we fine-tune the network with the consistency loss for additional 
$53k$ iterations.
Table~\ref{Table:sota_compare} summarizes the results for
the {\em hand-object network} and {\em hand-object network + consistency supervision} on all frames of the test set.
For the translation (T) aligned error, the root joint is the {\em wrist} joint.

Our {\em baseline} is the  hand branch of the {\em hand-object network} in~\cite{Hasson_2020_CVPR}.
Despite the training scheme in~\cite{Hasson_2020_CVPR}, we train our {\em baseline} longer for
$40k$ iterations with batch size $64$; we also augment the images with a random rotation in $[-\pi/8, \pi/8]$. We hypothesize that our longer training and rotation augmentation improved the baseline method compared to~\cite{Hasson_2020_CVPR}.

\end{document}